# Emotion-Oriented Behavior Model Using Deep Learning

Muhammad Arslan Raza, Muhammad Shoaib Farooq, Adel Khelifi, Atif Alvi

Department of Computer Science, School of System and Technology, University of Management and Technology, Lahore, 54000
Corresponding author: Muhammad Shoaib Farooq (shoaib.farooq@umt.edu.pk)

**Abstract:** Cognition has significance regarding social interaction among humans. In comparison, artificial agents are now becoming social partners of humans in different scenarios. Therefore, the interaction between artificial agents and humans must be adequate to maintain trust and increase believability. Emotions, as a fundamental ingredient of any social interaction, lead to behaviors that represent the effectiveness of the interaction through facial expressions and gestures in humans. Hence an agent must possess the social and cognitive abilities to understand human social parameters and behave accordingly. However, no such emotion-oriented behavior model is presented yet in the existing research. The emotion prediction may generate appropriate agents' behaviors for effective interaction using conversation modality. Considering the importance of emotions, and behaviors, for an agent's social interaction, an Emotion-based Behavior model is presented in this paper for Socio-cognitive artificial agents. The proposed model is implemented using tweets data trained on multiple models like Long Short-Term Memory (LSTM), Convolution Neural Network (CNN) and Bidirectional Encoder Representations from Transformers (BERT) for emotion prediction with an average accuracy of 92%, and 55% respectively. Further, using emotion predictions from CNN-LSTM, the behavior module responds using facial expressions and gestures using Behavioral Markup Language (BML). The accuracy of emotion-based behavior predictions is statistically validated using the 2-tailed Pearson correlation on the data collected from human users through questionnaires. Analysis shows that all emotion-based behaviors accurately depict human-like gestures and facial expressions based on the significant correlation at the 0.01 and 0.05 levels. This study is a steppingstone to a multi-faceted artificial agent interaction based on emotion-oriented behaviors. Cognition has significance regarding social interaction among humans.

## 1 Introduction

It is a challenge in artificial general intelligence (AGI) to enable emotional support in machines that perform intelligent and cognitive tasks. Such an artificial agent should have the ability to learn basic emotions from the conversation between an agent and humans and behave accordingly. In regular communication with humans, an agent must need emotion modelling for situation recognition, learn from the opponent's response, and respond appropriately [1]. With these abilities, agents could initiate conversations with other agents and humans. Emotions are negative and positive, and their presence offers implicit balance to cognitive processes and behavioral responses. An agent must be believable, trustworthy, and deliberative during the conversation. This study considers the basic emotions, i.e., Anticipation, Joy, trust, Fear, Surprise, Sadness, Disgust and Anger of Robert Plutchik's model [2]. Some studies used individual emotions for performance evaluation in different tasks; however, lacking the support of a complete set of basic emotions. For example, Anticipation, Joy [3], Trust [4], Fear, Surprise [5], [6], Sadness [7], Disgust [6], [7] and Anger [5] were used individually and sometimes partial groups were used.

Different modalities of conversation contain emotions and require a specific actual behavioral response. Textual modality is essential to focus on as an ingredient of social communication and data availability. Moreover, reinforcement, supervised learning, and unsupervised learning strategies are commonly used to train and evaluate textual modality [8], mainly text classification tasks based on best-suited machine learning algorithms [9].

Predicted emotions from the textual data lead to the mapping of behaviors for effective response. Textual prediction of emotions accompanied by gestural representations of behaviors makes the agent's interaction social and affective. Like emotions, behaviors also affect the cognitive process giving the essence of the socio-cognitive process behind the agent's interaction [10]. Therefore, Non- verbal behaviors such as gestural actions are required by the artificial agent to express emotions.

A model is proposed and tested in this research for emotion-oriented behavior that takes human input in textual form and predicts one of its basic emotions. Further, it maps the predicted emotion to non-verbal behavioral response bridging the gap between the agent and the humans, proving to be a trustworthy and believable partner in social interaction with humans.

Following are the contributions of this research work.

1. The proposed model predicts the agent's emotion against each interaction with a human or against events (see tab.2) to make it more cognitive and socially believable.
2. The agent's behaviors (sixteen) are derived from underlying emotions (eight) for the agent's self and another person/agent/event/object. (See tab. 3),
3. The emotionally derived agent's behaviors are simulated using BML Realizer to make agents present behaviors (gestures/facial expressions) like the human behaviors in socio-cognitive aspects
4. Dataset consolidation is performed as a single dataset (sentence level) with basic emotions (eight) labelling is not available with equal distribution. (See tab. 4)
5. The experiments are performed on a consolidated dataset, using seven different classifiers and four types of vectorizers for emotion prediction. (See tab. 6 for classifier comparisons)
6. Due to the variety of classes in the consolidated dataset and to compare results, the standard evaluation metrics of precision, recall, and f1 score are used.
7. From the artificial general intelligence (AGI) perspective, the proposed model leads an agent to assist humans in real-time scenarios through enhanced emotional stability and understanding.

The rest of the paper is structured as follows. Section 2 presents the related work regarding existing agent architectures and models to identify the research gap and need for emotion-based behavior modelling using eight emotions from the text. Moreover, existing techniques for emotion extraction and machine learning techniques are also reported. Section 3 presents the proposed model for emotion-based behavior and details of model training using an emotion-labelled dataset. Moreover, non-verbal behaviors are mapped to the emotion identified. Section 4 covers the implementation details, machine learning, and deep learning techniques used for emotion modelling. The dataset selection and preprocessing are also presented in the implementation section. Section 5 presents the implementation and testing results of emotion-based behavior models. The comparison of results using machine learning (ML) and Deep Learning, along with the statistical validation of emotions and behaviors, is also presented in this section. Section 6 presents the conclusion supporting the proposed model's requirement and novelty. Section 7 covers the limitations and future directions.

**2 Related Work**

This section briefly highlights the related works in emotion-based behavior modelling using machine learning.

The cognitive-affective architecture reported emotional beliefs, intentions, and desires to trigger emotions to help the agent in reasoning process. The study showed lack of validation and evaluation techniques to validate emotions and their associated behaviors. Moreover, complex phenomenon was needed to understand the modalities of emotions. However, these lack behavior modelling on an emotional basis. Emotional triggering leads to new goals and strategies and creates confident choices of behaviors and learning. Therefore, emotion-triggering parameters are needed to evaluate emotional models for socio-cognitive agents [10]. Biologically inspired agent architectures such as Emotional Biologically Inspired Cognitive Architecture (BICA) and Neuromodulating Cognitive Architecture (NEUCOGAR) enabled believability, behavior, learning and human collaboration at the abstract level lacking specific and detailed implementations. Hence, a specific and detailed behavioral model has been implemented in this research [11]. Emotions affect cognitive models such as the cognitive imagination model from the perspective of previously stored emotional memories and states. Conversely, the said QuBIC sub-architecture lacked in emotion and behavior based experiments. Thus, an emotion based behavioral approach is experimented [12].

The emotion model for the driver's situation shows an increase in emotional persistence with parallel display of similar states, improved efficiency, and goal orientation; however, the embedding of eight basic emotions in intelligent agents still needs improvement [13]. EBDI architecture also proposes using only primary and secondary emotions at the design level. However, this research presents a practical approach to modulate secondary emotions [14]. The simultaneous existence of emotions proposes a dimensional point of view, just like human's experience multiple emotions in different situations. ML is used to extract emotions and sentiments from the text, for example, the multi-classification method. Yet, this research proposed a categorical and cognitive point of view [15]. This study reported an overview of emotional systems and environments and stated that no agent architecture existed which implemented complex emotions for behavior generation. Such research encourages understanding the dynamics of emotions for artificial agents [16].

The existing architectures used specific modalities for communication, such as texts, videos, audio, or combinations. However, the motivation to use textual data as input for emotion appraisal is developed from current progress in Natural Language Processing (NLP) [17-20]. Emotion appraisal is required for applications such as dialogue-oriented and conversation systems [20] that lack the feature of emotion detection. Researchers reported different ML techniques for text classification, spam detection and malware identification [21]. the Ensemble Classifiers approach [22], and artificial neural networks (ANN) for data classification [17]. A bidirectional LSTM-CNN model for sentiment classification using parts of speech (POS) Tagging is proposed by Kumar [18]. Unfortunately, these need more data using eight primary emotions, as per Plutchik's model [2].

LSTM organizes the learnt features residing in textual data, remembers differences and similarities in data and improves with new input [19]. Moreover, CNN uses multiple filter techniques to detect and learn selected features from text patterns used for an open-domain question-answering system [20]. An academic emotion classification method for online learning using CNN and LSTM is also reported in [23]. Word embeddings define vectors to help train textual features of data [24]. Human Electroencephalogram (EEG) signals also depict emotional data using deep CNNs [25]. The BERT model has also reported emotion classification for eleven emotions for code-mixed text [26]. Detailed analysis of ANN shows that the recurrent nature of LSTM neural networks and layered filtering features of CNN algorithms are adequate for the current studies. However, no behavior was modelled in existing researches. Therefore, a combined algorithmic approach is adopted for modelling emotion-oriented behaviors.

A study involving simulation systems has inspired the usage of appraisal variables in emotion-based behavior modelling [27]. Such variables may be mapped on numerical or symbolic scales [9]. The contribution of varying basic and complex emotions with appraisal mechanisms is needed for human-like social interaction. Innate motivations fueled by implicit emotions form behavior-rendering agents' interactive abilities, and extraneous reactions improve behaviors [27]. The proposed study focuses on emotion representation in behaviors that must consider the variation in emotion appraisal variables. Some models suggest techniques and methods to derive and detect behaviors of humans for the implementation of similar behaviors in artificial agents based on emotional dynamics. For example, understanding the mental activities of a driver in a cognitive simulator during stop-and-seek behavioral activities with anticipation [22]. The emotional states affecting the behavioral responses of an artificial agent in a human-like manner based on real-time simulation and physical agents are reported. Like Hobbit and Pepper representing three emotions and their behaviors using five different modalities while interacting with real humans [27], [28]. Therefore, after evaluating and analyzing previous studies the proposed approach maps other behaviors as well

The validation study recorded behavior videos of both agents and presented them to online participants and evaluated the ideal distribution of agents' expressions. It gave a no significant p-value of $p>0.01$ in a Chi-square test where expression modality achieved the highest, that is, 87% from Hobbit and 50% from Pepper as compared to other less accurate modalities [28]. However, emotions change with scenarios and events, and so do behaviors. Universal behaviors are mapped from different psychological and computational sources, such as the given description of behavior for Joy. Varshavskaya [6] modelled behavior for Surprise. Jain et al. [5] gave a fuzzy model of behavior based on anger emotion. Behaviors related to trust-aware agents are also discussed [4]. Hudlicka [3] presented the behavior of sadness. In line with the research presented, eight primary emotions are modelled in this work, and their respective behaviors are mapped for artificial agents to perform better in human assistive scenarios. The following Tab. 1 summarizes the contributions and limitations of the related work in terms of modality/context used, the number of emotions considered, the kind of behaviors simulated, and the methodology used. The latest research needs to include dual behavior generation from specific eight emotions appraisal perspectives. Secondly, textual modality still needs to be reported for emotion-oriented behavior modelling. The current study uses textual data for eight emotions prediction using deep learning and agent's behavior mapping for all eight emotions using a BML realizer.

**Table 1:** Related work

| Ref | Modality/Context | Emotions | Behaviors | Method |
|---|---|---|---|---|
| [6] | Concept net | (2) Surprise, Disgust | Startle | Framework |
| [7] | Image | (2) Sadness, Disgust | Enthusiastic, Accept, Repel | Framework |
| [9] | No Modality | (5) Happy, Fear, Anger, Sad, hope | Defend, Depart | Framework |
| [23] | Academic context | (1) Anticipation | Stop & seek | Machine Learning |
| [27] | Robo Cup Rescue Simulation System (RCRSS) | (6) Anticipation, Joy, Fear, Sad, Disgust, Anger | Stop & seek/Move towards & move away | Simulation |
| **Current Study** | Text Data (sentence level) | (8) Anticipation, Trust, Joy, Surprise, Fear, Sad, Disgust, Anger | Enthusiastic, Accept, Retain, Startle, Defend, Regret, Depart, Hate | Deep Learning |

## 3 Proposed Model

As shown in Fig 1 below, the proposed emotion-oriented behavior model consists of multiple modules, including emotion classification, emotion appraisal, behavior generation and memory modules. The emotion appraisal module works as reported in [16], and memory modules (long-term and short-term) operate to memorize and learn emotion and behavior-specific data fetched and retained when an agent interacts with the environment. The interactive social environment provides the stimulus to the agent, and emotions are classified and appraised for that stimulus. These appraised emotions lead to behavior generation and actions (gestures) the agent performs towards the environment. A complete flow diagram of the model is presented in Fig. 1.

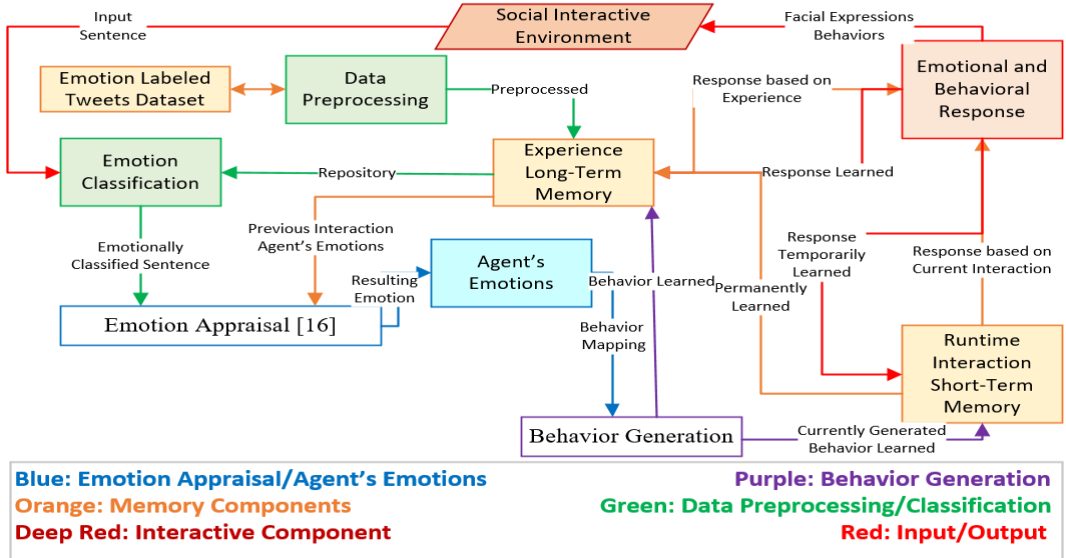

Figure: 1 Emotion-oriented Behavior Generation Model for Socio-Cognitive Agents

Emotion appraisal occurs when emotions compete or cooperate, resulting in one dominant emotion in the proposed model. Such emotion appraisal leads to modelling dual behaviors against each predicted emotion. Eight emotions are chosen from the second layer of Plutchik's model [2] for classifying textual data. Behaviors are modelled in terms of facial expressions and gestures to visually represent the dynamics of emotions and behaviors using natural language input for the artificial, and results of learning and prediction are presented.

The model begins with a single-sided conversation between a human and an artificial agent. Input is a sentence in the English language from a human user. In contrast, the artificial agent represents the output as facial expressions and simulated behaviors. Eight emotions are classified, with one of the eight labelled to one whole sentence, and its emotional context is also evident. Classification of emotions on deep learning-based trained models has enabled precise prediction of emotions in a sentence. The appraisal process of emotions [16] involves emotion dynamics, i.e., intensity and valence affect the resultant emotion.

The artificial agent to be addressed is socio-cognitive; therefore, predicted emotion against each interaction with a human provides the agent with its own emotions, as detailed in Tab. 2. These emotions are appraised, leading to agent behaviors. Positive emotions lead to positive events, and negative emotions lead to adverse events. Inspiration for similar emotions leading to behaviors has been taken from various sources [27], [28]. Interaction and response become the experience and can be used for future reference in long-term memory. Emotions, once predicted, are learned and kept for reference [30].

**Table 2:** Agent's emotions based on events

| Human's Emotion | Event (E-motivational Goal) | Agent's (Event-based) Emotion | Valence |
|---|---|---|---|
| **Anticipation** | Make Happen [27] | Anticipation, Hope [27] | Positive |
| **Trust** | Support [23] | Pity [23], Trust | Positive |
| **Joy** | Safe, Sustain [27] | Joy [27] | Positive |
| **Surprise** | The Unknown | Surprise | Neutral |
| **Fear** | Get to safety, Prevent [27] | Fear [27] | Negative |
| **Sadness** | Terminate, Getaway [27] | Distress [27], Sadness | Negative |
| **Disgust** | Dissociate [27] | Dislike [27], Disgust | Negative |

| Anger | Damage, Disappointment [27] | Anger | Negative |

After emotion prediction, the proposed model follows the derivation of behaviors from each predicted emotion. Tab. 3 presents all behaviors derived from the agent's eight emotions under study for the agent's self and another object that could be an agent or human. Agent's behaviors derived from the agent's emotions are recorded for the next interaction along with the text reference

**Table 3:** Agent's emotion-based behaviors for self and another person/object

| Agent's Emotion | Agent's Goal-based Behavior | Agent's Self Behavior | Agent's Other Behavior |
|---|---|---|---|
| **Anticipation, Hope [27]** | Find a family member Anticipate, Approach [27] | Enthusiastic [7] | Seek [9] |
| **Pity [23], Trust** | Help a person [23] | Accept [7], Rely [4] | Help [9] |
| **Joy [27]** | Jump up, Celebrate [27] | Retain | Affiliate [7] |
| **Surprise** | Nothing, Undefined | Startle [6] | Examine [9] |
| **Fear [27]** | Escape the risk, keep safe, get saved, Vigilance, Inhibition or Flight (run) [27] | Defend [9], Protect | Escape [7] |
| **Distress [27], Sadness** | Move around, Leave [27] | Regret [3] | Ignore [7] |
| **Dislike [27], Disgust** | Withdraw, Conceal, Submit [27] | Depart [9], Repel [7] | Avoid [7] |
| **Anger** | Fight, Quarrel | Hate [5] | Approach and Attack [9] |

## 4 Implementation of the Proposed Model

The implementation of the proposed study model follows detail of system and platform, i*mplementation* flow of the proposed model, dataset description for short texts, dataset quantity effecting precision, Vectorization and optimization and behaviors visualization.

### *4.1 System and Platform Details*

The specifications for conducting this research are a collection of programming languages with specific libraries used for training and algorithm implementation. Details are stated as under: -

- Python 3.6.4: Anaconda Environment: Jupyter Notebook
- Behavioral Markup Language: BML Realizer (SmartBody Simulator)
- Libraries: Keras 2.1.5 and 2.1.6, Pandas, NLTK, Scikit-learn 0.19.0, Pickle and TensorFlow (GPU)
- Pretrained GloVe word vectors on tweets (200 Dimensional)
- OS: Windows 10 Education 64-bit
- Processor: Intel Core i7-7500U CPU 2.70GHz 2.90GHz
- Installed Memory: 16 GB (15.9 GB usable)
- GPU support NVIDIA GeForce 940MX 2GB dedicated Secondary Memory: 30GB for all Project Files.

### *4.2 Implementation Flow of the Proposed Model*

In training an artificial agent for human-like emotional and behavioral abilities, some machine learning techniques (supervised, unsupervised and reinforcement) are applied to real-time human emotions and behaviors data. Fig. 2. Represents the module-wise implementation of the proposed model. It depicts the use of natural language processing techniques for input preprocessing, emotion classification using deep learning techniques, behavior generation, and visualization using a smart body simulator.

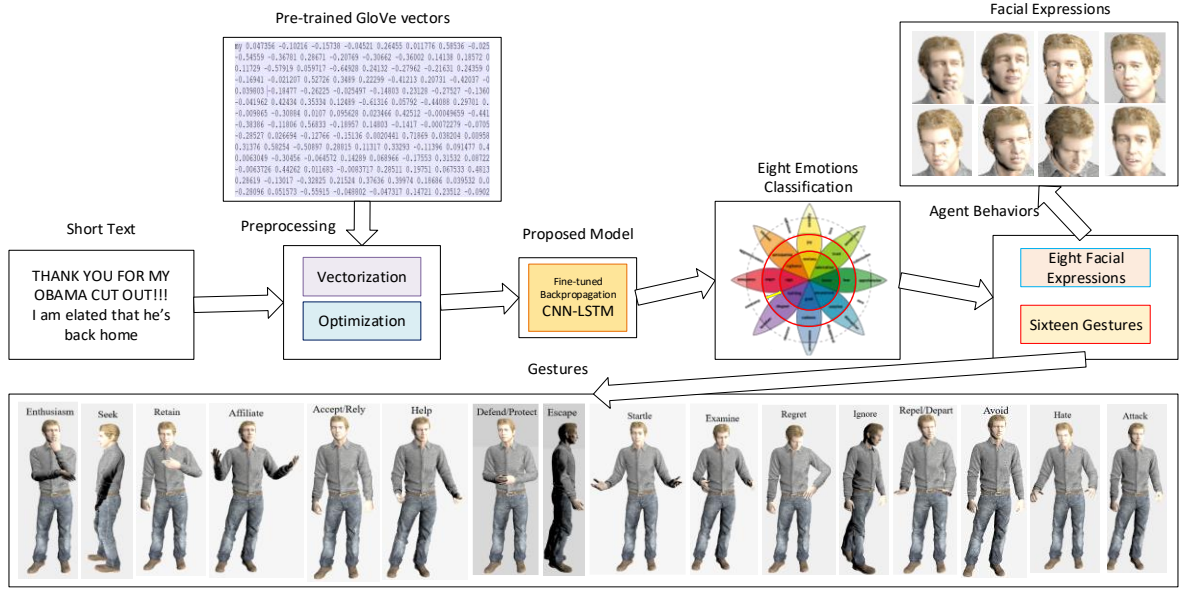

**Figure: 2** Module-wise implementations of the proposed model using the LSTM-CNN mode

The proposed model starts working with an input of short text sentence vectorized using pre-trained Global Vectors for Word Representation (GloVe). The optimizer selects the best-suited features from vectors, and the Adadelta optimizer is used. After this, a fine-tuned backpropagation LSTM-CNN model is used with a four-layer sequence, a convolution layer, a max pooling layer, a dense layer with Rectified Linear Units (ReLU) activation function, and a dropout layer with Softmax activation function to predict emotions. The predicted emotions are mapped to behaviors and display relevant facial expressions/gestures.

### *4.3 Dataset Description for Short Texts*

Data collection is the initial step toward emotion training, prediction, and behavior mapping. Collected data in Tab. 4 consists of published datasets from available sources containing labelled sentences of natural language (English) with eight emotions. These English sentences are random posts, comments etc., from Twitter. The combined dataset is divided into training, validation, and test sets. Before training, it is split into training and testing sets. The training set is further split into a validation set for sampling and results with a 2% validation split. The test set psychologically validates the new predicted emotions, and each sentence of the test set is evaluated through a trained model. The Natural Language Toolkit (NLTK) library preprocesses textual data by performing tokenization, lemmatization/stemming and stop-word removal for deep learning applications. GloVe vectors are used as retrained vectors during sentence preprocessing. These word embeddings represent most dimensions of various words in terms of difference [29-30]. To avoid retraining word vectors for commonly occurring words in one sentence, "GloVe vectors" are used. Training a model and then testing it requires classification techniques.

**Table 4:** Detailed selection of datasets and instances for processing

| S. N. | Dataset | Published | Emotions | Instances |
|---|---|---|---|---|
| **1** | National Research Council (NRC)-Tweet Emotion Intensity Dataset [31] | Yes | 4; Joy, Sadness, Fear, and Anger | **3868** |
| **2** | Sentiment Analysis: Emotion in Text [31] | No | 5; Joy, Sadness, Surprise, Anger, and Trust (Modified – Unknown and other positive Tweets filtered based on Trust lexicons) | **1698** **1000**(*Trust*) |
| **3** | Primary Emotions of Statements [31] | No | 4; Anticipation, Surprise, Disgust, and Trust | **1283** |

| 4 | Jan9-2012-tweets-clean.txt/GitHub repository [32] | No | 1; Disgust | **761** |
|---|---|---|---|---|
| All instances are combined in a single joint dataset. | | | | **8610** |

*4.4 Dataset Quantity Effecting Precision*

A dataset of four labelled emotions (i.e., anger, fear, sadness, and joy) is used in training and testing as a first step. Secondly, the subsequent four labelled emotion sentences (i.e., Surprise, disgust, anticipation, and trust) are added to the same dataset for training and testing. However, the Initial four emotions labelled sentences are abundant from various published sources leading towards better training and precise prediction results with an average precision of 76% of 3,613 instances, 57% of 18,943 instances, 53% of 596 instances, and 66% of 8,406 instances, separately. The subsequent four emotions labelled sentences are fewer than the initial four datasets. An equal number of sentences for each emotion label was required; therefore, further published sources of word-level emotion-labelled datasets were searched. Unfortunately, no other published datasets were found that met the criteria. Fortunately, all eight emotions for the word-level dataset are available; however, they are unavailable with similar quantities for sentence-level datasets [32]. Therefore, word-level datasets with required emotions are selected to extract the relevant sentences. A complete dataset with an equal number of instances is then trained on ML algorithms and tested. Published datasets with sentence-level emotion labelling are under development in ongoing research [33].

*4.5 Vectorization and Optimization*

Inspiration from existing classification techniques leads to training models using CNN and LSTM-CNN approaches [20][23][34]. Following the CNN approach, a multi-channel process is initially followed for emotion classification. Multi-channel refers to comparing two vector sets, both sets are trained with one kept static, and another is fine-tuned via backpropagation. It is a layered approach, and techniques are applied in a sequence for desired output. Filters applied to both sets provided features. Optimization of parameters is done through the Adadelta update rule as it requires lesser training epochs than other optimizers; however, it gives similar results as the Adagrad rule. This technique provides a pool of features. Applying the Max-pooling technique on a pool of extracted features gives resultant features with a maximum value. Multiple resultant features are then treated with a fully connected Softmax technique for the probability distribution of emotion labels. Dropout restriction is applied on the outputs with l2-normalization scheme on weight vectors to avoid beneficial values from randomly dropping out, using Regularization. For vectorizing, GloVe pre-trained vectors are used with an embedding dimension of 200. The second approach involves two phases of implementation. The LSTM algorithm is combined with the CNN approach. Therefore, forming the first phase a CNN-LSTM application and, later, an LSTM-CNN application on the same dataset. In the first phase, the CNN layer takes pre-trained word embeddings to extract features and then feeds these features to the LSTM layer to learn text sequences. The second phase applies a reverse approach with the LSTM-CNN approach. It identifies and learns the features extracted from word embeddings, and then these learned features are pooled out to a dimension smaller than the extracted one. CNN-LSTM gives better results in comparison to LSTM-CNN. Existing approaches identified the polarity of labelled tweets; however, the multi-class classification of eight emotions primarily focuses on the current research. A detail of parameters is shown in Tab. 5, and a pictorial representation of model implementation is represented in Fig.3

**Table 5:** Parameters for fine-tuning and analysis of the effect

| **Parameters** | Embedding Dimension | Epoch | Batch Size | Kernel Size | Pool Size | Drop out | Word Embedding's |
|---|---|---|---|---|---|---|---|
| CNN-LSTM | 200 [29] | 400 | 128 [35] | 2, 3, 5, 6, 8 (Conv1) [30] | 4 [35] | 0.5 [35] | Pre-trained (GloVe) |

|                    |                |
|--------------------|----------------|
| and 4, 3, 2        | twitter-27B    |
| (Conv2) [35]       | [29]           |

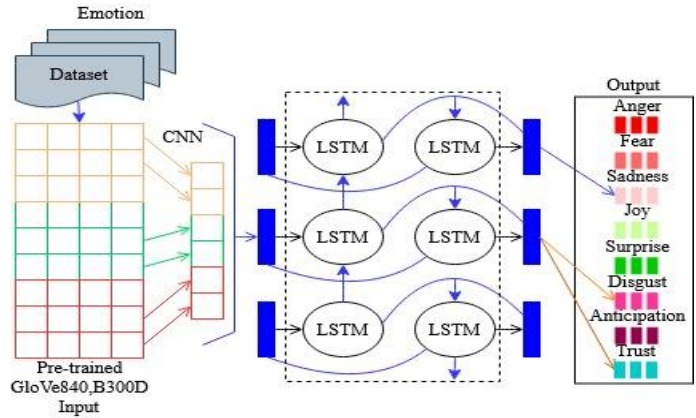

**Figure: 3** CNN-LSTM Model for eight basic emotions classification

In comparison to CNN-LSTM, the BERT base case model is also applied to the same dataset which provided an accuracy of 55%. The hyperparameters for BERT model implementation include 16 batch size, 2e-5 learning rate, 10 epochs and AdamW for optimization. The CNN-LSTM provided better accuracy than the BERT model.

*4.6 Behaviors' Visualization*

Behaviors are displayed in gestures and facial expressions based on predicted emotion. Each behavior represents an action or event occurring when a specific emotional stimulus is triggered. Depending on the prediction, each emotion leads to dual specific behaviors from the agent's perspective under consideration. Section 5.2 below shows behaviors generated using a simulator application as a realizer to BML. Facial expressions represent emotions, and gestures represent the behaviors in this part of the proposed model. Current behaviors are recorded in short-term memory. Trained models, previously predicted emotions, and derived behaviors are recorded in long-term memory.

**5 Results and Validations**

There are three experiments; the first consists of test results of emotion prediction from a classifier trained on the CNN-LSTM algorithm and formulating an agent's emotions. It also includes a comparison with six other models. The second part consists of behavior generation from the classifier output in the form of gestures using a simulation. The third part consists of a statistical analysis produced to validate the items in the dataset and validate the agent's emotions along with its formed and visualized behaviors.

*5.1 Classifiers Comparison*

The first part of the experiment starts with training a model of eight labelled emotions on a mix of classifiers, models, and vectorizers. BERT model, CNN-LSTM classifier and three other classifiers, such as Stochastic Gradient Decent (SGD) classifier, Multi-Layer Perceptron (MLP) classifier, and Linear Support Vector Classifier (SVC), is used for training. Among the SGD classifiers are the Logistic Regression model and the linear model. The logistic regression model is fitted with three varying vectorizers: Bag-of-words, Term frequency – Inverse document frequency (Tf-IDF), and Hashing. In contrast, the Linear model is only fitted with one vectorizer, Bag-of-words. The MLP and Linear SVC classifiers used the linear and Support Vector Machine (SVM) models, respectively, with a Bag-of-words vectorizer for both. Comparison results are presented in Tab. 6. It is presented that the CNN-LSTM classifier performed better than the other classifiers on the given dataset. Afterwards, the emotions are classified in the validation and test set. This classification of emotions formulates the emotions in artificial agents. The time complexities of AI models have their effect on training and classification. Based on the used models in Tab. 6, the time complexity of Logistic Regression, Linear Model, and Layered Model is O(n). However, the SVM Model has a time complexity of O(n3).

Table 6: Classifiers comparison with CNN-LSTM classifier

| Classifier | Model | Vectorizer | Average Precision |
|---|---|---|---|
| SGD Classifier | Logistic Regression | Bag-of-words | 0.51 |
| SGD Classifier | Logistic Regression | Tf-idf | 0.50 |
| SGD Classifier | Logistic Regression | Hashing | 0.46 |
| MLP Classifier | Logistic Regression | Bag-of-words | 0.46 |
| SGD Classifier | Linear Model | Bag-of-words | 0.50 |
| Linear SVC | SVM Model | Bag-of-words | 0.50 |
| CNN-LSTM | Layered Model | pretrained GloVe Vectors | 0.92 |
| BERT | Layered Model | Transformer | 0.55 |

*5.1.1 Evaluation Metrics*

Standard evaluation metrics of precision, recall and F1 score are considered for this study and reported in the literature [36-38]. The classification report is shown in Tab. 7 with each emotion predicted in terms of evaluation metrics. An average of 0.92 precision is obtained from the CNN-LSTM classifier and dataset. Firstly, these metrics are selected due to the variety of classes in a combined dataset. Secondly, the comparison of varying values is based on their likelihood in the data. A confusion matrix is generated for the CNN-LSTM classifier after the classification of validation sets, as shown in Fig. 3.

Table 7: Standard evaluation metrics for CNN-LSTM classifier

|  | CNN-LSTM | | | | BERT | | | |
|---|---|---|---|---|---|---|---|---|
|  | Precision | Recall | F1-score | Support | Precision | Recall | F1-score | Support |
| **Joy** | 0.96 | 0.92 | 0.94 | 239 | 0.56 | 0.73 | 0.63 | 41 |
| **Sadness** | 0.91 | 0.91 | 0.91 | 221 | 0.53 | 0.54 | 0.53 | 57 |
| **Fear** | 0.93 | 0.94 | 0.93 | 245 | 0.68 | 0.71 | 0.70 | 17 |
| **Anger** | 0.94 | 0.91 | 0.92 | 206 | 0.75 | 0.70 | 0.72 | 54 |
| **Surprise** | 0.86 | 0.94 | 0.90 | 226 | 0.33 | 0.28 | 0.31 | 60 |
| **Anticipation** | 0.85 | 0.91 | 0.88 | 182 | 0.40 | 0.32 | 0.35 | 53 |
| **Disgust** | 0.95 | 0.90 | 0.93 | 192 | 0.50 | 0.59 | 0.54 | 41 |
| **Trust** | 0.97 | 0.94 | 0.96 | 196 | 0.70 | 0.65 | 0.67 | 48 |
| **Average** | 0.92 | 0.92 | 0.92 | 1707 | 0.55 | 0.56 | 0.56 | 431 |

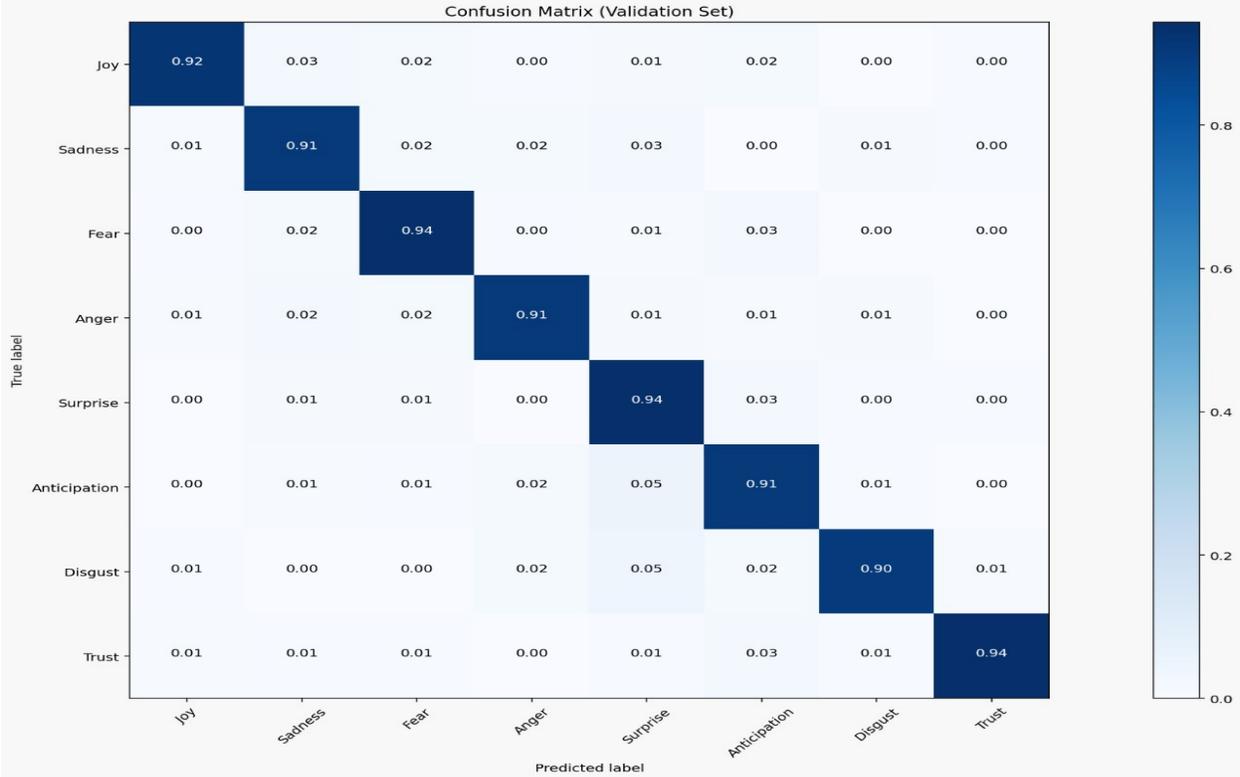

**Figure 4:** Confusion matrix of the validation set with eight basic emotions

*5.1.2 Classifying the Test Set*

Specific parameters affect precision, such as epochs. Training results are produced at 400 epochs. Sentences for each emotion are tested individually on the trained model. Ten test sentences from each emotion are taken. It is observed in the results shown below that falsely predicted sentences are closely related to the original emotion as per psychology. For example, when testing sentences for the prediction of joy, some false predictions are also positive emotions such as anticipation and Surprise; however, one true negative is fear which is oddly related to human feelings in that sentence. These results show 65% precision correctly tested on test sentences. A few difficulties exist, such as the validation of the dataset. Emotion-labelled sentences' datasets of eight basic emotions are fewer and contain unequal instances for each emotion. Some researchers are working on creating such datasets. This study combines existing datasets into one dataset with an equal number of instances against each emotion and names it the common dataset. Fig. 5 shows five sentences from a test set predicted for the emotion of joy. Similar predictions are made on the other seven emotions as well.

```
Bridget Jones' Baby was bloody hilarious
Prediction: Fear

happy birthday jin young!!!!!!
Prediction: Anticipation

Watching football matches without commentary is something that I rejoice, found a transmission of City match like that today, j
oyful.
Prediction: Joy

comes on soon
Prediction: Surprise

THANK YOU FOR MY OBAMA CUT OUT!!!!!! I am elated that he's back home
Prediction: Joy
```

**Figure 5:** A test case for the prediction of emotion: joy

## 5.2 Behaviors Formulation

Artificial agents' emotions and behaviors are formed and visualized in the second part of the experiment. The dataset and the formulated emotions and behaviors are analyzed and validated using statistical techniques in the third part of the experiment. Figures below show behaviors generated from the simulator application as a realizer to BML, Namely, Fig. 6 & Fig. 7.

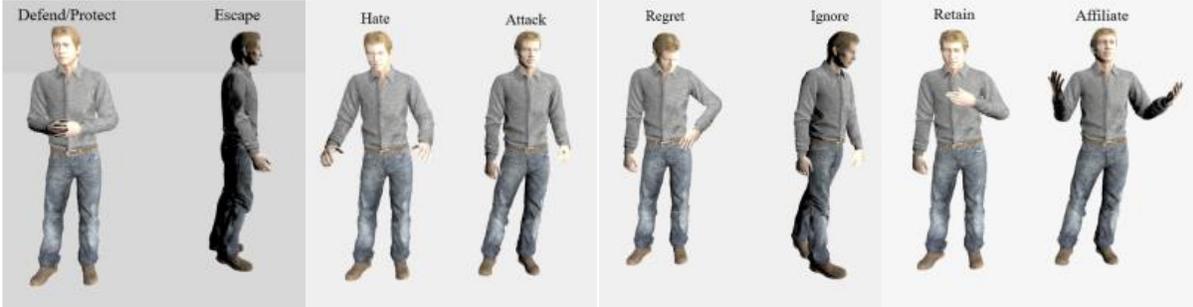

**Figure 6:** Behaviors representation for (a) fear (b) anger (c) sadness and (d) joy emotions using BML

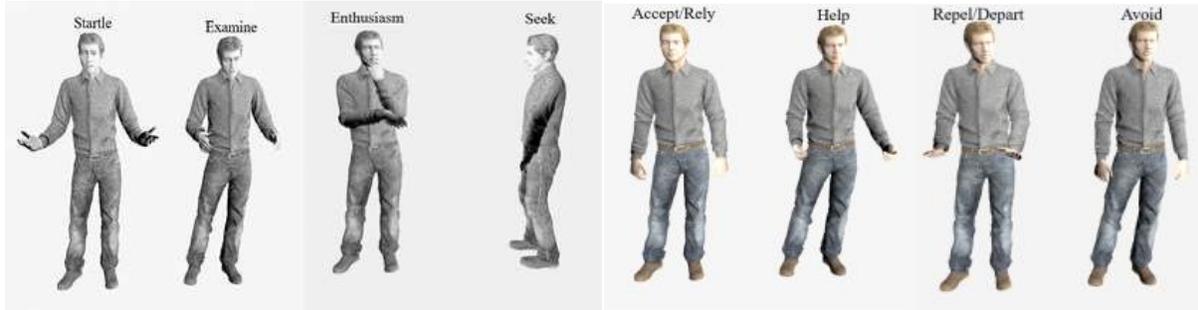

**Figure 7:** Behaviors representation for (a) surprise (b) anticipation (c) trust and (d) disgust using BML

## 5.3 Statistical Validation of Human Predictions for Emotions

Statistical validation of data provides the authenticity of labelled emotions and predicted behaviors. Participants from six educational institutions (students and faculty) in Lahore City attempted a questionnaire of 80 sentences, each representing a single emotion out of the primary eight under study. A total of 400 participants accurately filled out the questionnaire. Tab. 8 shows the participant-to-institution ratio, and Tab. 9 shows participants who correctly identified each emotion under study. Ten sentences were purposefully extracted from the dataset to be tested with human prediction. A total of 80 sentences were selected and termed as items ranged A1-A80 in this study. Predicted emotions are joy, anger, disgust, trust, Surprise, sadness, anticipation, and fear.

Using the Pearson correlation method, two-tailed significance has validated all selected items in Tab. 9. A sum score calculates all items as correlated. P-values are calculated to represent significant correlation values of $p<0.05$ and $p<0.01$, as shown in Tab. 10, generated by statistical tools. There are eight correlation tables, each representing data from one of eight emotions. Therefore, the data taken from the dataset and the questionnaires are valid and ensure the authenticity of emotions predicted per item by humans.

**Table 8:** Institution-wise distribution of participants

| Institution | UET | NCBA&E | UCP | LGU | UMT | UOL |
|---|---|---|---|---|---|---|
| **Frequency** | 100 | 51 | 40 | 60 | 70 | 79 |
| **Valid %** | 25.0 | 12.8 | 10.0 | 15.0 | 17.5 | 19.8 |

**Table 9:** Correctly identified emotions by 400 participants.

| Items | A1-10 | A11-20 | A21-30 | A31-40 | A41-50 | A51-60 | A61-70 | A71-80 |
|---|---|---|---|---|---|---|---|---|
| **Emotion** | Anger | Fear | Sadness | Joy | Surprise | Disgust | Anticipation | Trust |
| **Correct Predict** | 82.81% | 86.61% | 86.14% | 93.92% | 90.79% | 89.36% | 89.99% | 84.75% |
| **Avg. Freq.** | 331.1 | 346.3 | 344.4 | 375.6 | 63.1 | 357.2 | 359.8 | 338.9 |

**Table 10:** Correlation scores of eight emotions showing the validity of data.

|  | Anticipation | Joy | Trust | Surprise | Fear | Sadness | Disgust | Anger |
|---|---|---|---|---|---|---|---|---|
| Pearson Correlation | 0.130 | 0.190 | 0.407 | 0.127 | 0.452 | 0.538 | 0.116 | 0.459 |
| Sig. (2-tailed) | 0.024 | 0.001 | 0.000 | 0.011 | 0.000 | 0.000 | 0.045 | 0.000 |
| N | 300 | 300 | 400 | 400 | 400 | 400 | 300 | 400 |

*5.4 Behavior Analysis*

After emotion analysis, behavior analysis is necessary. The second questionnaire validates each behavior against eight emotions. Participants were randomly selected for behavior analysis. A total of 50 participants, as shown in Tab. 11, observed sixteen different behaviors on agent simulation while emotions were not disclosed for these behaviors. Participants were then asked to predict each emotion represented by shown behaviors. Interviews conducted after the display of simulated agents showed that few participants predicted slightly wrong emotions. However, most participants judged correctly, as in Tab. 12. Unique representation of behaviors against each emotion prevents correlation calculation among behavior items.

**Table 11:** Interviews conducted in institutions.

| **Institution** | LGU | NCBA&E | UCP | UET | Total |
|---|---|---|---|---|---|
| **Frequency** | 15 | 5 | 15 | 15 | 50 |
| **Valid %** | 30.0 | 10.0 | 30 | 30 | 100 |

**Table 12:** Correctly identified behaviors

| **Emotion** | Anger | Fear | Sadness | Joy | Surprise | Disgust | Anticipation | Trust |
|---|---|---|---|---|---|---|---|---|
| **Correct Pred. %** | 86% | 92% | 96% | 92% | 96% | 96% | 98% | 90% |
| **Wrong Pred. %** | 14% | 8% | 4% | 8% | 4% | 4% | 2% | 10% |

After emotion analysis, behavior analysis is performed. The second questionnaire validates each behavior against eight emotions. Participants were randomly selected for behavior analysis. A total of 50 participants, as shown in Tab. 11, observed sixteen different behaviors on agent simulation while emotions were not disclosed for these behaviors. Participants were then asked to predict each emotion represented by shown behaviors. Interviews conducted after the display of simulated agents showed that few participants predicted slightly wrong emotions. However, most participants judged correctly, as in Tab. 12. Unique representation of behaviors against each emotion prevents correlation calculation among behavior items.

## 6 Conclusion

An emotion-based model is presented that leads to non-verbal behaviors for socio-cognitive agents to maintain the essence of social interaction. Eight basic emotions are predicted from the textual input based on the trained classifier of combined CNN-LSTM algorithms. The dynamics of emotions are satisfied in sixteen different behavioral representations and eight specific facial expressions. Textual prediction supported by visual representations enhanced understandability and trust between the agent and the human.

A comparative study is conducted with real-time human prediction to validate the labelled training set and analyze visually represented non-verbal behaviors. Pearson correlation is calculated for validating the emotion-labelled dataset, and the p-value came out to be significant where p<0.01 and p<0.05. Behavior analysis shows that all representations are reliable and valid per undergraduate and graduate human individuals at academic institutes in Pakistan.

## 7 Limitations and Future Work

Currently, this research is based on textual data labelled with eight emotions only; however, it may be improved in future by identifying values of eight appraisal variables for each emotion following the componential approach. It might improve precision in emotion prediction and allow the modelling of complex emotions (tertiary level in Plutchik Wheel) and real-time behaviors for socio-cognitive agents. Secondly, the current model could be implemented using visual modality, i.e., images or facial expressions. Moreover, complex emotion-oriented behavior modelling would provide more believable assistive robots for two-way conversation in human-agent social scenarios.

**Acknowledgement:** Thanks to families and colleagues who provided us with moral support.

**Funding Statement:** No funding was received for this research.

**Conflicts of Interest:** The authors declare they have no conflicts of interest to report regarding the present study.